\documentclass[10pt,a4paper]{article}

\usepackage[T1]{fontenc}
\usepackage[margin=18mm]{geometry}
\usepackage{amsmath,amssymb}
\usepackage{graphicx}
\usepackage{booktabs}
\usepackage{makecell}
\usepackage{multirow}
\usepackage{cite}
\usepackage[hyphens]{url}
\usepackage{hyperref}
\usepackage{orcidlink}
\usepackage{cleveref}
\usepackage[font=small,skip=3pt]{caption}
\usepackage[compact]{titlesec}

\titlespacing*{\section}{0pt}{1.5ex plus .4ex minus .2ex}{.7ex}
\titlespacing*{\subsection}{0pt}{1.2ex plus .3ex minus .2ex}{.5ex}
\titlespacing*{\paragraph}{0pt}{.8ex plus .2ex minus .1ex}{.7em}

\title{LAION-Mobile: Evaluating Deepfake Detectors On One Million Smartphone Photos}
\author{Achim von Stryk\orcidlink{0009-0004-7479-5747}\\
  \small Stralsund University, Germany\\
  \small \href{mailto:achim.vonStryk@gmail.com}{\nolinkurl{achim.vonStryk@gmail.com}}
  \and
  Janis Keuper\orcidlink{0000-0002-1327-1243}\\
  \small IMLA, Offenburg University, Germany\\
  \small \href{mailto:keuper@imla.ai}{\nolinkurl{keuper@imla.ai}}}
\date{}

\begin{document}
\setlength{\abovedisplayshortskip}{4pt plus 2pt minus 2pt}
\setlength{\belowdisplayshortskip}{4pt plus 2pt minus 2pt}
\maketitle
\vspace{-2.5em}

\begin{abstract}
Most Deepfake detectors report near-perfect AUC scores on their reference benchmarks. However, a
recent ICML position paper argues that these evaluations collectively neglect the impact of modern smartphone photography:
the widely used on-device neural image-signal processing pipelines (like multi-sensor fusion or noise and
motion-blur suppression) increasingly shift the imaging paradigm from simple lens projections towards computational photography.  
Hence, devices actually \emph{generate}, rather than record photos. This increases the risk that deepfake detectors may flag ordinary phone photos as fake. Due to the lack of large-scale datasets containing images from modern smartphones, this hypothesis has so far only been tested in small proof-of-concept studies. The aim of this paper is to close this gap.\\
We introduce ``LAION-Mobile'', an open dataset containing $\sim 1$ million smartphone images with EXIF metadata distilled from re-LAION-5B. Evaluating twelve state-of-the-art deepfake detectors with their original paper
checkpoints on a $9{,}115$-image evaluation sample of this pool (DIRE on $738$), we report three
key findings: (i)~On modern AI content no detector exceeds AUC $0.624$, and five
of twelve fall below chance. (ii)~Real-photo false-alarm rates are an artefact
of threshold calibration: thresholds fitted on legacy GAN data make several detectors look deployable ($\le$$11\%$ FPR), yet the same detectors flag $17$--$91\%$ of
real photos once the identical criterion is refit on modern content.
(iii)~Consequently, no detector both beats chance on modern AI content and
keeps a deployable real-photo false-alarm rate. Mirroring the device mix of
web collections, the corpus probes the \emph{first} neural-ISP generation
($2018$--$2020$); current flagships are essentially absent, leaving the
modern-ISP regime as the open gap.

The ``LAION-Mobile'' dataset, including per-image EXIF metadata is available at \url{https://huggingface.co/datasets/laionmobile/laion-mobile}.
\end{abstract}

\section{Introduction}
\label{sec:intro}

More than $90\%$ of photographs today are taken with
smartphones~\cite{woolf2025mobilestats}. This makes them 
the most important data source for the evaluation of image-based algorithms. However,
in the context of deepfake detection, a recent ICML
position paper~\cite{keuper2026real} showed that current benchmarks collectively neglect 
smartphone images in their evaluation and instead rely on older internet sources for "real" image samples.
Consequently, \cite{keuper2026real} argues that this is highly problematic, since modern smartphones heavily rely
on computational photography: On-device neural
image-signal-processing (ISP) stacks denoising, motion blur suppression
and exposure burst fusion into a single high-dynamic-range
image, using learned models conceptually close to generative
networks~\cite{masud2026authenticity}; these are
just some of the neural algorithms marketed as Apple Smart HDR, Google
HDR+~\cite{hasinoff2016hdrplus} or Samsung Scene Optimizer. Hence, these computational-photography pipelines increasingly \emph{generate}
rather than merely record images. In consequence, the line between a ``real'' photograph and a
synthetic one erodes, and a detector trained to flag ``generated'' content
risks flagging ordinary camera captures as fake.

So far, this argument has been made in ~\cite{keuper2026real} from first principles and demonstrated only on a small, controlled dataset; whether it holds at the scale and diversity of real
smartphone photography remained mostly open. 
This paper aims to close this gap and to verify the key concerns raised in~\cite{keuper2026real}.

Modern deepfake detectors report detection AUC $\ge 0.99$ on their
reference benchmarks~\cite{wang2020cnndetection,ojha2023univfd,li2025iapl},
but on data that closely resemble their own training distribution. We instead
audit twelve detectors (eleven published methods plus a DINOv3 variant of
RIGID) with their \emph{original paper checkpoints} and measure how often they
flag real, unmodified photographs from a re-LAION-5B-derived pool of
$935{,}399$ smartphone images (\cref{sec:laion_mobile}) with on-device ISP processing already baked in. For context we also evaluate on the NTIRE 2026
challenge benchmark~\cite{gushchin2026ntire}, a modern AI-generated mixture.
Cross-generator shift is the main failure mode already
discussed in depth in the literature~\cite{wang2020cnndetection,yan2024aide}; it serves here as a
calibrated \emph{ordinary} distribution shift, against which smartphone
photographs prove the harder case: the same detectors that merely degrade on
NTIRE collapse into flagging the majority of real phone photos as fake. This
large-scale result confirms the central concern of the position
paper~\cite{keuper2026real}.

\paragraph{Contributions.}
(i)~We assemble a publicly-derivable real-photo evaluation subset of $9{,}115$
smartphone images from the LAION-Mobile re-LAION-5B collection ($\sim$1\,M
EXIF-identified IDs) and release the filter manifest and per-image EXIF
(\cref{sec:laion_mobile}), turning the position paper's
claim~\cite{keuper2026real} into a reusable measurement instrument.
(ii)~We audit twelve detectors with original checkpoints on three regimes:
in-domain ForenSynths (correctness check), NTIRE 2026 (modern AI mix), and the
LAION-Mobile real photos (\cref{sec:results}).
(iii)~We show that the low false-alarm rates several detectors appear to enjoy
are an artefact of the calibration set: a threshold fitted on legacy GAN data
understates the real-world rate several-fold, and once it is fitted on
modern AI content, \emph{no} detector both beats chance there and stays
deployable on real photos (\cref{sec:calibration}).

\begin{figure}[tbp]
  \centering
  \includegraphics[width=\linewidth]{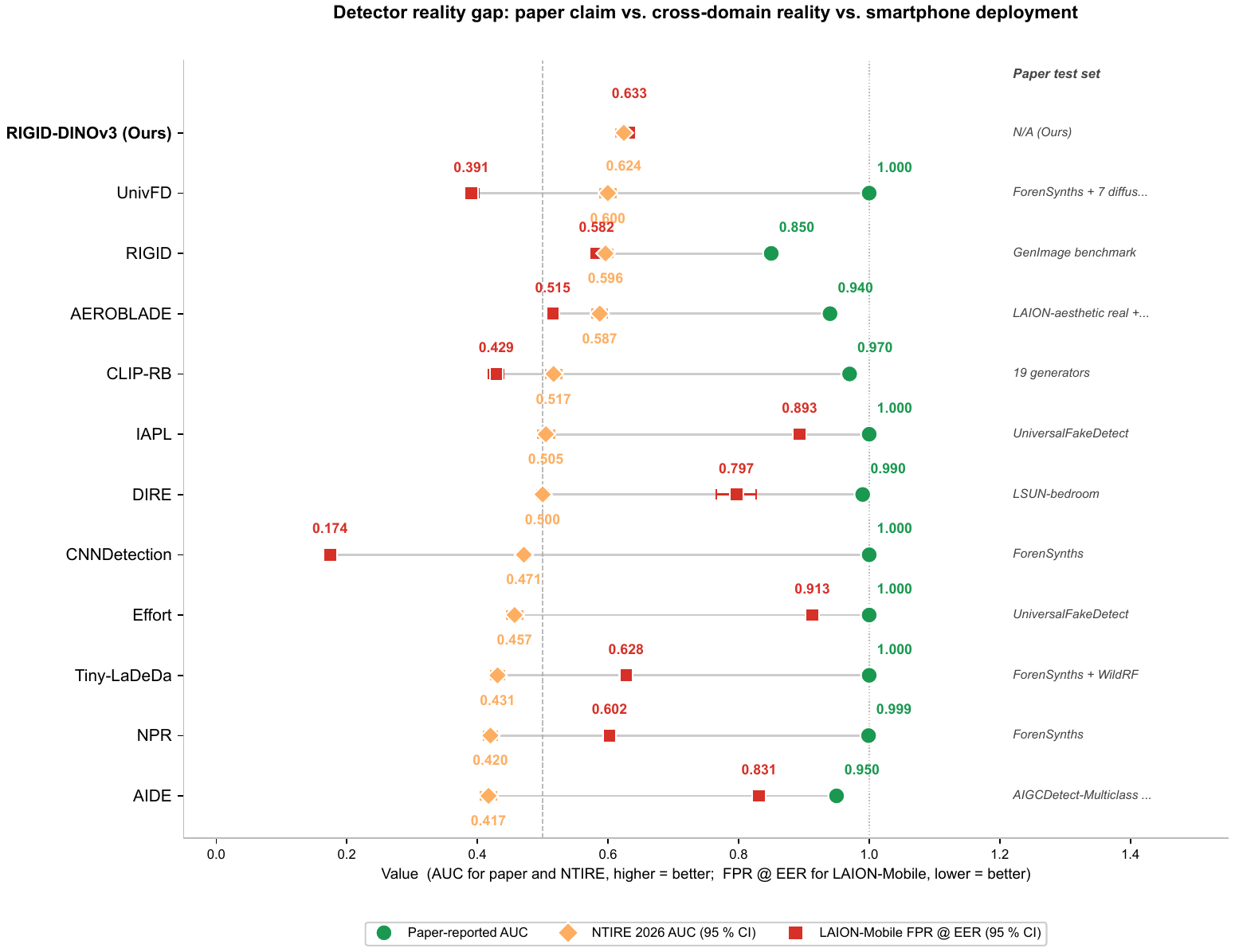}
  \caption{The deepfake-detection reality gap. For each detector, we visualize three different measurements: paper-reported AUC on its own test set (green); NTIRE 2026
  AUC~\cite{gushchin2026ntire} (orange, $95\%$ bootstrap CI); and the
  LAION-Mobile false-alarm rate at the detector's own NTIRE-calibrated EER
  threshold (red, $n{=}9{,}115$, DIRE on $738$; the $\tau(\textrm{NTIRE})$
  column of \cref{tab:laion_mobile-fpr}). Paper-AUC bars are published
  headlines under each method's own protocol (context only); directly
  comparable are the NTIRE and LAION-Mobile columns under our uniform protocol
  (\cref{sec:replication}). A deployable detector needs a \emph{high} NTIRE
  AUC \emph{and} a \emph{low} LAION-Mobile FPR; across the roster the two are
  mutually exclusive. Results are sorted by descending NTIRE AUC.}
  \label{fig:reality-gap}
\end{figure}

\Cref{fig:reality-gap} summarizes the result: near-perfect paper AUC coexists
with NTIRE AUC no higher than $0.624$ and, at modern-calibrated
operating points, LAION-Mobile raises false-alarm rates up to $91\%$.

\section{Related Work}
\label{sec:related_work}

\paragraph{Deepfake detection methods.}
Modern methods divide along three lines: (i) Convolutional fingerprint detectors
utilize the spectral artefacts left by the up-sampling layers of generative
networks, which fail to reproduce natural spectral
distributions~\cite{durall2020watch}. Prominent examples for this approach are
CNNDetection~\cite{wang2020cnndetection},
NPR~\cite{tan2024npr}, Tiny-LaDeDa~\cite{cavia2024ladeda}.
(ii) CLIP~\cite{radford2021clip}-feature linear probes exploit semantic
representations (UnivFD~\cite{ojha2023univfd},
CLIP-Raising-the-Bar~\cite{cozzolino2024cliprb},
IAPL~\cite{li2025iapl}, Effort~\cite{yan2024effort}).\\
(iii) Reconstruction-error and training-free representation-invariance detectors form
the third family (DIRE~\cite{wang2023dire},
AEROBLADE~\cite{ricker2024aeroblade}, RIGID~\cite{he2024rigid});\\
AIDE~\cite{yan2024aide} hybridises DCT-band selection with a
ConvNeXt-XXL classifier. We include representatives of each family (twelve
detectors, counting our variant that swaps RIGID's
DINOv2~\cite{oquab2024dinov2} backbone for DINOv3~\cite{simeoni2025dinov3}).

\paragraph{Cross-domain evaluation.}
Brittleness across generators is well documented:
Wang \emph{et al.}~\cite{wang2020cnndetection} already report drops from $1.000$ on
ProGAN~\cite{karras2018progan} to $0.66$ on DeepFake, and the ``sanity check''
work~\cite{yan2024aide} shows similar overfitting in newer detectors.
What this leaves open is the position paper's claim
itself~\cite{keuper2026real}: Existing real-photo evaluations are too small to
tell whether computational-photography artefacts make detectors misfire at
scale: they are curated laboratory captures (SIDD~\cite{abdelhamed2018sidd};
Google's HDR+ burst set~\cite{hasinoff2016hdrplus}) of at most $10^3$ images
from a handful of devices. Hence, we introduce the LAION-Mobile subset
(\cref{sec:laion_mobile}) to close this gap, extending real-photo evaluation by an
order of magnitude in size and device diversity. Unlike challenge mixes such as
NTIRE~2026~\cite{gushchin2026ntire}, which pair \emph{general} real and
generated imagery and score AUC, LAION-Mobile is real-only and phone-native, isolating the
false-alarm rate on unmodified consumer photographs.

\section{The LAION-Mobile Smartphone-Photo Subset}
\label{sec:laion_mobile}

\begin{figure}[tbp]
  \centering
  \includegraphics[width=0.9\linewidth]{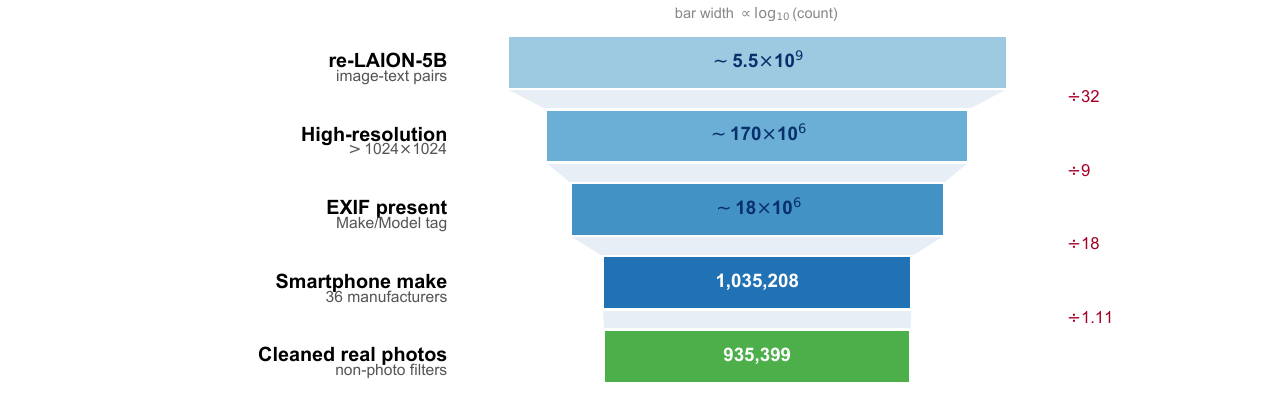}
  \caption{Distillation funnel from re-LAION-5B to the cleaned LAION-Mobile
  pool. Each tier is annotated with its image count and the reduction factor
  from the stage above; bar width is proportional to
  $\log_{10}(\text{count})$, as the five stages span four orders of magnitude.
  The green tier is the $935{,}399$-image cleaned pool from which all
  evaluation images are drawn (non-photo filtering detailed below).}
  \label{fig:funnel}
\end{figure}

\paragraph{Source and filtering.}
In order to validate the hypothesis proposed in~\cite{keuper2026real},
we build a real-photo set from publicly available smartphone images. Starting from
re-LAION-5B~\cite{relaion2024}, the safety-reviewed 2024 re-release of
LAION-5B~\cite{schuhmann2022laion5b}, we narrow the image samples in three steps
(\cref{fig:funnel}): (i)~the high-resolution subset (over
$1024\times1024$)~\cite{laion2022highres}; (ii)~those carrying EXIF, for which
re-LAION distributes \texttt{Make}/\texttt{Model}; (iii)~IDs whose EXIF make
matches one of $36$ smartphone manufacturers (Apple, Samsung, Xiaomi, \emph{etc.}).
After perceptual-hash/URL deduplication, dropping dead URLs and
manufacturer-balanced sampling, $1{,}035{,}208$ unique IDs remain (the
``LAION-Mobile-1M'' manifest).

\begin{figure}[tbp]
  \centering
  \includegraphics[width=0.88\linewidth]{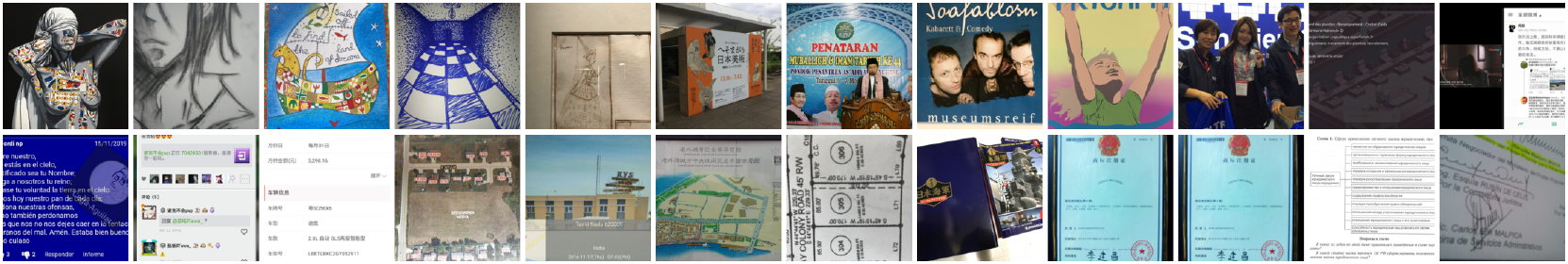}
  \caption{A sample of the $99{,}809$ entries removed by the three-stage
  non-photo filter, spanning its categories (in reading order): illustrations
  and paintings, logos and graphics, social-media screenshots, maps, and
  document scans. Each carries a smartphone \texttt{Make}/\texttt{Model} in its
  EXIF yet is not a photograph of a physical scene. Thumbnails for illustration
  only; imagery is not redistributed.}
  \label{fig:filtered-grid}
\end{figure}

\paragraph{Removing non-photographs.}
An EXIF \texttt{Make}/\texttt{Model} can be written by editing software, so a
match alone does not guarantee a raw camera capture. We therefore apply three
independent non-photo filters and discard the union of their flags.
\emph{(i)}~A caption filter regex-matches the per-image text annotations, flagging $69{,}407$ images ($6.7\%$) as illustrations ($44{,}270$), logos
and graphics ($13{,}985$), and the remainder UI screenshots, documents, maps,
QR codes and placeholders. \emph{(ii)}~An aspect-ratio filter flags the tall
display ratios ($\max/\min > 1.9$) typical of screenshots. \emph{(iii)}~An
EXIF-completeness filter rejects entries exposing \emph{none} of focal length,
aperture, exposure time or ISO. The heuristics are complementary: a screenshot
whose caption hallucinates a real scene is still caught by aspect ratio or
missing EXIF. The union flags $99{,}809$ entries, leaving $935{,}399$ cleaned
real-photo IDs, from which all evaluation images are drawn;
\cref{fig:filtered-grid} shows a sample of the removed content.

\begin{figure}[tbp]
  \centering
  \includegraphics[width=0.92\linewidth]{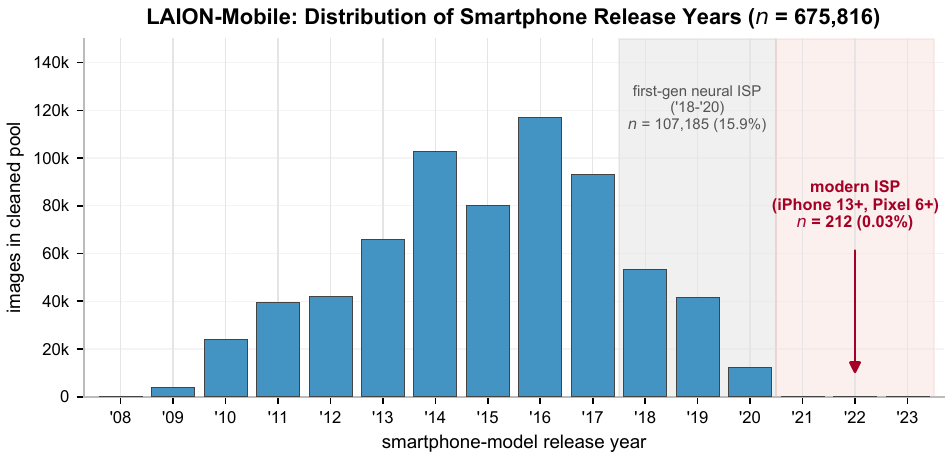}
  \caption{Device-era coverage of the \emph{full} cleaned LAION-Mobile pool
  ($935{,}399$ images): release-dated images per smartphone-model release year
  ($n{=}675{,}816$ release-dated; detector- and threshold-independent). The pool
  peaks at $2014$--$2016$ hardware and collapses after $2020$: only $212$
  images ($0.03\%$) come from phones released in $2021$ or later, so the
  modern-ISP regime (iPhone~13+, Pixel~6+) the position paper targets
  is essentially absent. The $2018$--$2020$ \texttt{heavy} bin of \cref{sec:ki-era}
  is only the \emph{first} neural-ISP generation.}
  \label{fig:device-coverage}
\end{figure}

\begin{figure}[tbp]
  \centering
  \includegraphics[width=0.9\linewidth]{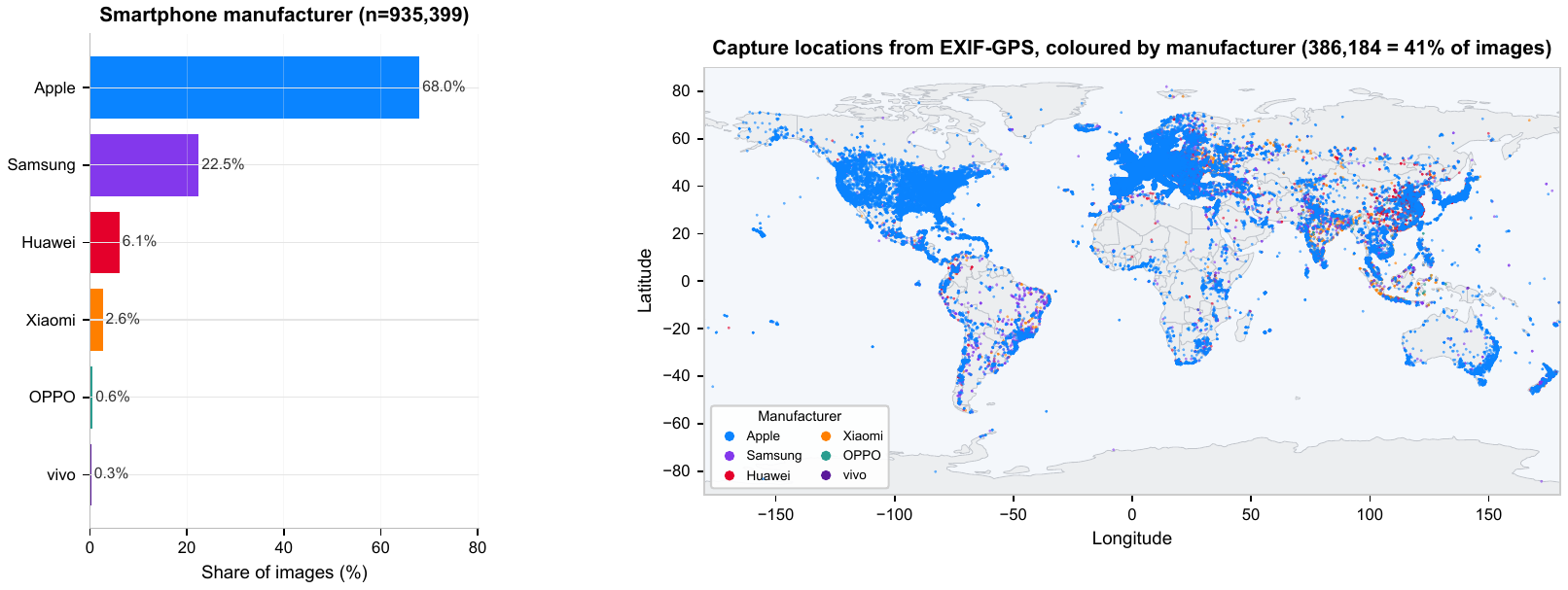}
  \caption{Provenance of the cleaned LAION-Mobile pool ($n{=}935{,}399$).
  \emph{Left:} share of images per smartphone manufacturer, each bar in its
  brand colour. \emph{Right:} EXIF-GPS capture locations of the geotagged
  images ($386{,}184$, $41.3\%$), each point coloured by the same
  manufacturer palette.}
  \label{fig:provenance}
\end{figure}

\begin{figure}[tbp]
  \centering
  \includegraphics[width=0.88\linewidth]{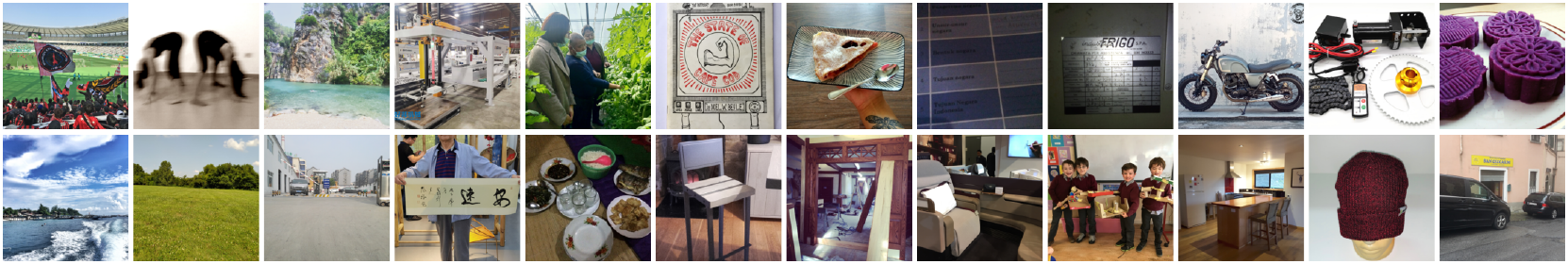}
  \caption{Random, uncurated sample of the cleaned LAION-Mobile real-photo pool
  ($24$ images, manufacturer-stratified; the pool holds $935{,}399$).
  Thumbnails for illustration only; imagery is not redistributed. The
  occasional residual non-photograph reflects the imperfect heuristic
  (\cref{sec:limitations}).}
  \label{fig:samples}
\end{figure}

\paragraph{Composition.}
The cleaned pool's manufacturer mix is
dominated by Apple ($68.0\%$) and Samsung ($22.5\%$), then Huawei ($6.1\%$),
Xiaomi ($2.6\%$), OPPO and vivo, reflecting the device composition of web image
collections rather than market share. The most frequent models (\cref{tab:phone_models}) are older
iPhones (iPhone~6, 6s, 7, 5s, 4S, 5), i.e.\ devices from roughly
$2011$ to $2017$ that largely predate neural ISPs (\cref{fig:device-coverage}); this skew is
relevant to \cref{sec:ki-era}, which consequently sees few heavily-processed
devices. Valid, in-range EXIF-GPS coordinates are available for 41.3\% of the cleaned
pool (\cref{fig:provenance}), concentrated in Europe, North America and East
Asia ($94.2\%$ northern-hemisphere); a few implausible ones ($18$ Antarctic
coordinates) confirm that \texttt{real} is a high-precision heuristic, not
GPS-verified ground truth (\cref{sec:limitations}). Apple and Samsung dominate every populated region, so brand and geography are entangled rather than independent (\cref{fig:provenance}). \Cref{fig:samples} shows an uncurated sample.

\begin{table}[tbp]
  \caption{The twelve most frequent phone models in the cleaned pool
  ($n{=}935{,}399$), with ISP era
  (\texttt{none}/\texttt{light}/\texttt{heavy} ML aggressiveness, classified
  from manufacturer specifications; cf.\ \cref{sec:ki-era}). The pool is
  dominated by pre-$2018$ devices: only $10.9\%$ of images come from
  \texttt{heavy}-ISP phones (iPhone~XS and later), against $40.9\%$
  \texttt{none}, $18.4\%$ \texttt{light} and $29.8\%$ unclassified. This skew
  motivates the future-work direction in \cref{sec:conclusion}.}
  \label{tab:phone_models}
  \centering
  \small
  \setlength{\tabcolsep}{6pt}
\begin{tabular}{@{}llrr@{}}
\toprule
Model & ISP era & Images & Share \\
\midrule
    iPhone 6 & none & 74\,294 & 7.9\% \\
    iPhone 6s & none & 51\,232 & 5.5\% \\
    iPhone 7 & light & 49\,035 & 5.2\% \\
    iPhone 5s & none & 47\,940 & 5.1\% \\
    iPhone 4S & none & 39\,644 & 4.2\% \\
    iPhone 5 & none & 39\,023 & 4.2\% \\
    iPhone 7 Plus & light & 33\,142 & 3.5\% \\
    iPhone X & light & 30\,146 & 3.2\% \\
    iPhone 4 & none & 24\,178 & 2.6\% \\
    iPhone 8 Plus & light & 23\,222 & 2.5\% \\
    iPhone 8 & light & 22\,570 & 2.4\% \\
    iPhone 6 Plus & none & 20\,817 & 2.2\% \\
\bottomrule
\end{tabular}

\end{table}

\paragraph{Evaluation subset.}
We download a manufacturer-balanced random subset of $10{,}000$ IDs; $9{,}171$
are retrieved ($\sim$$8\%$ URL rot) and $9{,}115$ carry valid detector scores.
Each image keeps its EXIF (\texttt{Model}, focal length, aperture, ISO,
exposure time, pixel dimensions); the set spans five major manufacturers,
$\sim$$80$ models and an exposure range of seven $f$-stops. \Cref{tab:datasets}
summarizes it alongside the other evaluation sets.

\Cref{tab:exif-completeness} reports per-field EXIF completeness over the
full pool: the funnel guarantees only that a \emph{row} carries an EXIF block
with a smartphone make, so individual fields may still be empty. The exposure
triangle (ISO, aperture, exposure time) and the capture timestamp are
populated for $98\%$ of images, making a physical illumination proxy
(the ISO-100 exposure value $\mathrm{EV}_{100}$) computable almost everywhere
(\cref{sec:ki-era}). Two intuitively interesting variables are \emph{not}
analysable: the standard ambient-temperature tag is absent from the corpus
entirely (phones write thermal data, if at all, only into the proprietary
binary \texttt{MakerNote}), and autofocus distance is populated for $0.3\%$
of images, since smartphones do not write their focus state back into
standard EXIF.

\begin{table}[tbp]
  \caption{Per-field EXIF completeness over the smartphone-make pool
  ($n{=}1{,}035{,}208$; the cleaned subset is drawn from these). Fields below
  $50\%$ (lower block) are excluded from analysis.}
  \label{tab:exif-completeness}
  \centering
  \small
  \setlength{\tabcolsep}{5pt}
  \begin{tabular}{@{}lr@{\hskip 18pt}lr@{}}
    \toprule
    EXIF field & \% & EXIF field & \% \\
    \midrule
    Make / Model            & $100.0$ & Scene/exposure modes$^{\dagger}$ & $95.0$--$96.8$ \\
    DateTimeOriginal        & $98.4$  & BrightnessValue                  & $90.6$ \\
    FocalLength             & $98.3$  & ExposureBias                     & $81.9$ \\
    ISO / FNumber / Exp.\ time & $97.8$--$98.2$ & MakerNote (binary)     & $58.9$ \\
    Flash                   & $97.6$  &                                  &        \\
    \midrule
    GPSInfo                 & $46.7$  & SubjectDistance (AF)             & $0.3$  \\
    LightSource             & $14.4$  & AmbientTemperature               & $0.0$  \\
    SubjectDistanceRange    & $5.6$   &                                  &        \\
    \bottomrule
  \end{tabular}
  \par\smallskip
  {\footnotesize $^{\dagger}$WhiteBalance, MeteringMode, ExposureProgram,
  SceneCaptureType, ExposureMode.}
\end{table}

\paragraph{What ``real'' means, and labels.}
We call an image \texttt{real} if it is a camera capture of a physical scene,
regardless of on-device ISP processing, and \texttt{fake} only if a generative
model synthesised its \emph{content}. A Smart HDR or Night Mode photograph is
therefore real, and a detector that flags it is in error: the audit asks
precisely whether detectors honour this content-level notion of authenticity or
instead react to the processing pipeline. All LAION-Mobile images are
\texttt{real} (label $=0$), so the set supports one-class evaluation
(false-positive rate) only; we obtain operating points from EER thresholds fit
on held-out two-class data (\cref{sec:calibration}).

\paragraph{Release and ethics.}
We release (a)~the cleaned LAION-Mobile~1M ID list with the three non-photo
flags, (b)~the EXIF-enriched metadata of the $9{,}115$-image sample with a
per-image content hash, and (c)~the per-image scores, so future detectors can
be added without re-downloading and the headline numbers stay reproducible as
URLs rot. Per LAION policy we redistribute manifests, metadata and scores only; we do
not redistribute imagery, and downstream users fetch images from the original
URLs.
Where licences permit we additionally provide a frozen archive of the
evaluation sample. We rely on the maintainers' content filtering and recommend
the safety-reviewed re-release for reproduction.

\begin{table}[tbp]
  \caption{Datasets used in this study; ``Real''/``Fake'' are the total images
  available in each dataset.}
  \label{tab:datasets}
  \centering
  \small
  \setlength{\tabcolsep}{6pt}
\resizebox{\columnwidth}{!}{%
\begin{tabular}{@{}llrrl@{}}
\toprule
Dataset & Source & Real & Fake & Used for \\
\midrule
\textbf{Paper test sets} & & & & \\
ForenSynths-13gen   & \cite{wang2020cnndetection}  & 45\,169 & 45\,160 & implementation replication \\
\midrule
\textbf{Modern AI mix} & & & & \\
NTIRE 2026 shard\,0 & \cite{gushchin2026ntire}     & 17\,982 & 32\,018 & modern cross-domain test \\
\midrule
\textbf{ISP-processed paired} & & & & \\
ProGAN-ISP          & ours, Neural-ISP on \cite{wang2020cnndetection} & 1\,689 & 1\,700 & ISP-effect study \\
\midrule
\textbf{Real-only smartphone / camera} & & & & \\
HDR+ subset         & \cite{hasinoff2016hdrplus}   & 153 bursts & n/a & smartphone ISP baseline \\
SIDD-Medium-sRGB    & \cite{abdelhamed2018sidd}    & 160 scenes & n/a & smartphone capture baseline \\
MIDD                & \cite{flepp2024midd} & 1\,732 & n/a & sensor-stratified baseline \\
\textbf{LAION-Mobile (this work)} & re-LAION-5B-filt.\ \cite{relaion2024} & \textbf{935\,399} & n/a & headline real-world test \\
\bottomrule
\end{tabular}%
}

\end{table}

\section{Methodology}
\label{sec:methodology}

\paragraph{Detectors and protocol.}
We evaluate the twelve detectors of \cref{tab:laion_mobile-fpr}, each with its
\emph{original paper checkpoint} where available; we never substitute a
fine-tuned variant when a paper checkpoint is published. CLIP-Raising-the-Bar
needs a configuration override (OpenAI CLIP backbone with QuickGELU instead of
the CommonPool default; inverted score convention) to match its training setup,
documented in the released code. Each forward pass uses the input size and normalisation of
the original paper (\emph{e.g.}, $224{\times}224$ for CLIP-family, $256{\times}256$ for
AEROBLADE / AIDE / DIRE). For two-class data (ForenSynths variants, NTIRE 2026)
we report AUC; for the one-class LAION-Mobile set we report FPR at the default
$\tau{=}0.5$ and at a per-detector EER threshold fit \emph{separately} on each
two-class set, $\tau(\textrm{ProGAN-ISP})$ and $\tau(\textrm{NTIRE})$. Reporting
both exposes \emph{calibration drift} (legacy GAN vs.\ modern AI). On two-class
data we additionally report TPR at fixed $1\%$/$5\%$ FPR budgets, the operating
points a deployment would fix. Throughout, \emph{deployable} means beating
chance on modern AI content while keeping a low single-digit real-photo
false-alarm rate at the calibrated operating point. Bootstrap
$95\%$ CIs use $2{,}000$ stratified resamples.

\paragraph{Coverage.}
We evaluate each detector on a fixed random sample of every dataset, so the CIs
reflect the evaluation sample size; \cref{tab:datasets} lists the total
available per source, which exceeds the evaluated sample (for LAION-Mobile, the
$9{,}115$ scored images of \cref{sec:laion_mobile}; DIRE on $738$, as its
per-image diffusion reconstruction is far costlier than a forward pass).

\section{Results}
\label{sec:results}

\subsection{Implementation Correctness (ForenSynths Replication)}
\label{sec:replication}

\begin{figure}[tbp]
  \centering
  \includegraphics[width=\linewidth]{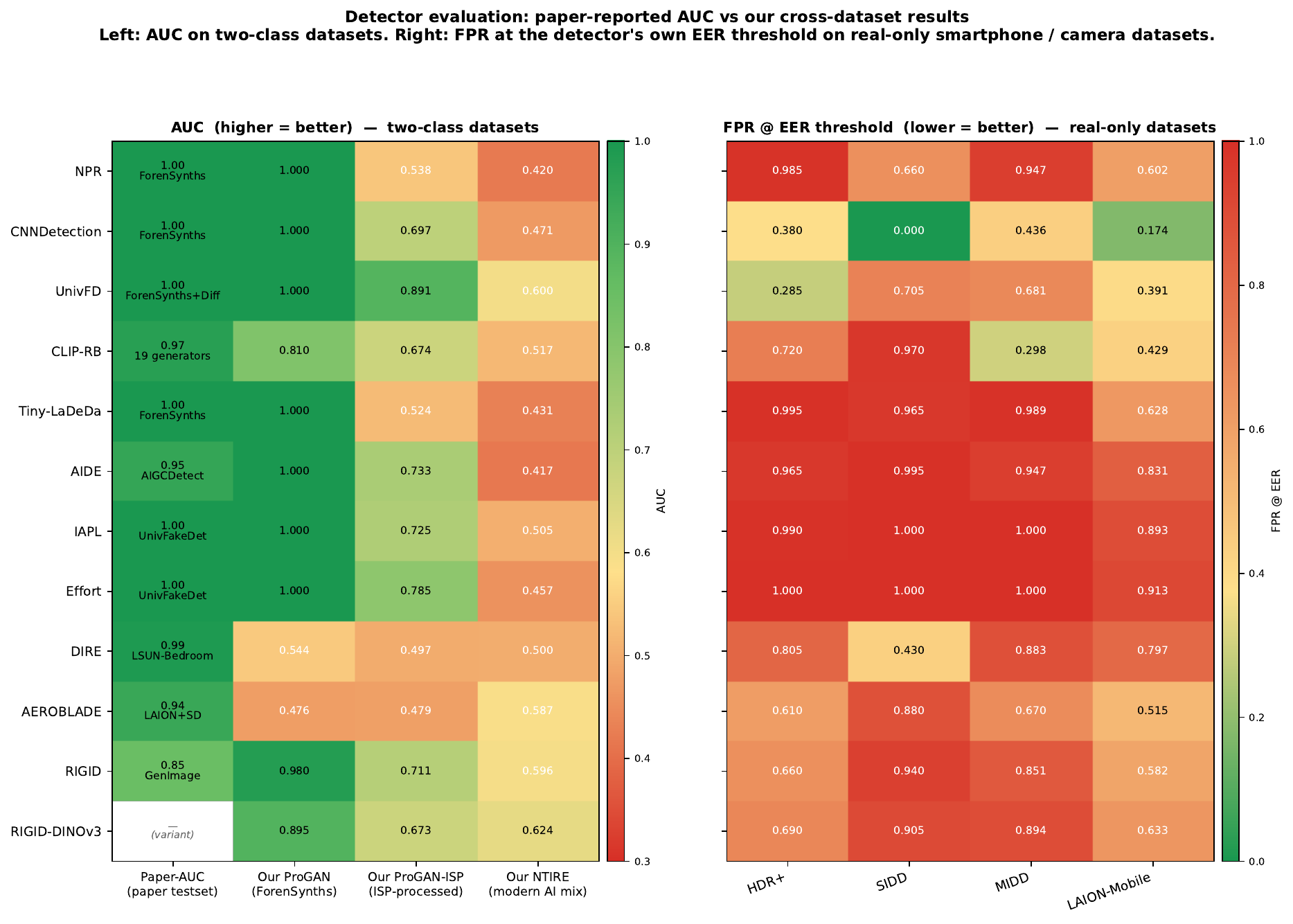}
  \caption{All-datasets matrix. \emph{Left:} AUC on two-class datasets (higher
  better): paper-reported, then our ProGAN-ISP and NTIRE 2026. \emph{Right:}
  false-positive rate on real-only smartphone/camera datasets at each detector's
  NTIRE-calibrated EER threshold (lower better). $\text{FPR}{=}0$ can reflect a
  discriminating detector or a default-biased one outputting \texttt{real}
  unconditionally; the left column disambiguates.}
  \label{fig:matrix}
\end{figure}

Using the original paper checkpoints on the original ForenSynths test
set~\cite{wang2020cnndetection} ($13$ generators, $n{=}200$ per class), the
three detectors with explicit per-generator paper tables reproduce the in-domain
ProGAN row within $0.001$ AUC, and $18$ of $25$ ($72\%$) per-generator cells
reproduce within $0.05$ (the largest gaps fall in categories the original papers
flag as hard). This confirms our architecture, preprocessing and weight-loading
are correct; detectors without published per-generator tables are sanity-checked
against their paper headline and our ProGAN AUC (\cref{fig:matrix}, ``Our
ProGAN'').

\subsection{Cross-Domain Failure on Modern AI Content}
\label{sec:ntire}

On the NTIRE 2026 shard~0 benchmark ($n{=}10{,}000$; $3{,}610$ real /
$6{,}390$ fake), NTIRE AUC ranges from $0.42$ to $0.62$, with $95\%$ bootstrap
CIs of about $\pm0.01$ (\cref{tab:laion_mobile-fpr}). Only five detectors are
statistically above chance and two are indistinguishable from it; the remaining
five sit \emph{below} $0.5$ with CIs excluding it, having inverted their score
ordering on modern content. Even the best detector clears chance by only
$0.12$ AUC. Fixed false-alarm budgets sharpen the picture further: no detector
exceeds $13.5\%$ TPR at $5\%$ FPR, or $3.8\%$ at $1\%$
(\cref{tab:laion_mobile-fpr}).

\subsection{False Alarms on Real Smartphone Photos}
\label{sec:laion_mobile-results}

NTIRE is an ordinary cross-generator shift on which detectors merely degrade;
real smartphone photos are the harder case. We report LAION-Mobile FPR under
two operating points (\cref{tab:laion_mobile-fpr}). At $\tau{=}0.5$ the rates are
uninformative: several detectors show $0.0\%$ while another flags $98.1\%$,
because the biased detectors output \texttt{real} regardless of input. The fair
reading is the FPR at each detector's EER threshold, but \emph{which} two-class
set fixes it decides the answer: calibrated on the legacy GAN set (ProGAN-ISP)
several detectors look deployable ($\le$$11\%$ FPR, UnivFD as low as $0.2\%$),
yet re-fitting the \emph{same} criterion on modern NTIRE content lifts every one
of them to $17$--$91\%$. This gap is calibration drift: the safe-looking
operating point is set on a 2019-era GAN distribution unrelated to anything a
deployed detector now sees. At the modern-calibrated threshold no detector
occupies the deployable quadrant of high NTIRE AUC and low LAION-Mobile FPR:
even the most deployable detector, UnivFD, still flags $39.1\%$, and the only one
under $20\%$ is itself below chance on NTIRE.

The same pattern holds on three independent curated real-photo sets (HDR+, SIDD,
MIDD; right panel of \cref{fig:matrix}): the high-FPR detectors misfire on the
large majority of these laboratory captures too, so the false alarms are not a
LAION artefact.

\begin{table*}[tbp]
  \caption{Per-detector summary on the three regimes. ``Paper AUC'' is the
  published headline; ``NTIRE AUC'' is on the modern challenge
  mix~\cite{gushchin2026ntire}, followed by TPR at fixed $1\%$/$5\%$ FPR
  budgets on the same set (\cref{sec:methodology}). The two ``LAION-Mobile
  FPR'' columns are the smartphone-photo false-positive rate at each
  detector's EER threshold, calibrated per dataset: $\tau(\textrm{ProGAN-ISP})$
  (legacy GAN) versus $\tau(\textrm{NTIRE})$ (modern). Their gap is the
  calibration drift. All $95\%$ bootstrap CIs in brackets; detectors sorted by
  NTIRE AUC.}
  \label{tab:laion_mobile-fpr}
  \centering
  \small
  \setlength{\tabcolsep}{5pt}
  \resizebox{\textwidth}{!}{%
  \begin{tabular}{@{}l r l l l l l@{}}
    \toprule
    Detector & Paper AUC & NTIRE AUC (95\% CI)
      & \shortstack[l]{NTIRE TPR\\@ $1\%$ FPR}
      & \shortstack[l]{NTIRE TPR\\@ $5\%$ FPR}
      & \shortstack[l]{LAION-Mobile FPR\\@ $\tau$(ProGAN-ISP) (95\% CI)}
      & \shortstack[l]{LAION-Mobile FPR\\@ $\tau$(NTIRE) (95\% CI)} \\
    \midrule
RIGID-DINOv3 & NA & $0.624\ [0.612,\ 0.635]$ & $0.017$ & $0.065$ & $0.977\ [0.974,\ 0.980]$ & $0.633\ [0.623,\ 0.643]$ \\
UnivFD & $1.000$ & $0.600\ [0.588,\ 0.611]$ & $0.029$ & $0.103$ & $0.002\ [0.001,\ 0.003]$ & $0.391\ [0.380,\ 0.401]$ \\
RIGID & $0.850$ & $0.596\ [0.584,\ 0.607]$ & $0.018$ & $0.084$ & $0.979\ [0.977,\ 0.982]$ & $0.582\ [0.573,\ 0.592]$ \\
AEROBLADE & $0.940$ & $0.587\ [0.576,\ 0.599]$ & $0.010$ & $0.068$ & $0.946\ [0.941,\ 0.950]$ & $0.515\ [0.505,\ 0.525]$ \\
CLIP-RB & $0.970$ & $0.517\ [0.505,\ 0.528]$ & $0.015$ & $0.062$ & $0.061\ [0.056,\ 0.066]$ & $0.429\ [0.418,\ 0.439]$ \\
IAPL & $1.000$ & $0.505\ [0.493,\ 0.517]$ & $0.038$ & $0.135$ & $0.021\ [0.018,\ 0.024]$ & $0.893\ [0.887,\ 0.900]$ \\
DIRE & $0.990$ & $0.500\ [0.492,\ 0.508]$ & $0.000$ & $0.000$ & $0.797\ [0.766,\ 0.827]$ & $0.797\ [0.766,\ 0.827]$ \\
CNNDetection & $1.000$ & $0.471\ [0.464,\ 0.478]$ & $0.008$ & $0.029$ & $0.105\ [0.099,\ 0.111]$ & $0.174\ [0.166,\ 0.182]$ \\
Effort & $1.000$ & $0.457\ [0.446,\ 0.469]$ & $0.022$ & $0.081$ & $0.036\ [0.032,\ 0.040]$ & $0.913\ [0.907,\ 0.919]$ \\
Tiny-LaDeDa & $1.000$ & $0.431\ [0.420,\ 0.440]$ & $0.000$ & $0.032$ & $0.384\ [0.374,\ 0.395]$ & $0.628\ [0.618,\ 0.638]$ \\
NPR & $0.999$ & $0.420\ [0.410,\ 0.430]$ & $0.000$ & $0.034$ & $0.466\ [0.456,\ 0.476]$ & $0.602\ [0.593,\ 0.612]$ \\
AIDE & $0.950$ & $0.417\ [0.405,\ 0.429]$ & $0.005$ & $0.028$ & $0.446\ [0.436,\ 0.457]$ & $0.831\ [0.823,\ 0.839]$
\\
    \midrule 
    \multicolumn{7}{l}{\footnotesize At $\tau{=}0.5$ the FPR is bimodal: $0.0\%$ for real-biased detectors, up to $98.1\%$ (DIRE).} \\
    \bottomrule
  \end{tabular}}
\end{table*}

\subsection{Computational-Photography Era: the Modern Regime is Missing}
\label{sec:ki-era}

Since real smartphone images and AI-generated images now share a family of
neural-network artefacts, the position paper~\cite{keuper2026real} predicts
that phones with heavier neural ISPs produce \emph{higher} false-alarm rates. Labelling
each phone model \texttt{none}/\texttt{light}/\texttt{heavy} from its
manufacturer's pipeline description and comparing the \texttt{heavy} and
\texttt{none} bins per detector (two-proportion $z$-test, Benjamini--Hochberg
corrected), we find the prediction does not hold: only two detectors move with
it, four move significantly \emph{against} it (most strongly Tiny-LaDeDa,
$\Delta{=}-0.494$, and AIDE, $-0.388$), and the remaining six are
indistinguishable from $\Delta{=}0$.

Reading this inversion as ``modern ISPs are harmless'' would be wrong; it is an
artefact of \emph{which} phones a web corpus contains. Every \texttt{heavy}
ISP image comes from a $2018$--$2020$ device (iPhone~XS through
iPhone~12, Galaxy~S9): the \emph{first} generation of default-on neural ISPs,
dominated by classical multi-frame \emph{denoising} rather than the
generative texture synthesis of current flagships. Along a continuous,
threshold-free axis (the Spearman correlation between a detector's fake-score
and the phone-model release year over $2010$--$2020$), the high-frequency
detectors decline monotonically (e.g.\ Tiny-LaDeDa $\rho{=}-0.40$, AIDE
$-0.34$; all $p{<}10^{-70}$), consistent with newer ISPs \emph{removing} the
noise residue these detectors mistake for synthesis rather than adding
generative artefacts. Crucially, the corpus stops exactly where the hypothesis
becomes interesting (\cref{fig:device-coverage}): almost none of its
release-dated images come from phones released in $2021$ or later, and none from
the iPhone~13--15 or Pixel~6+ class the position paper targets. Even among
images \emph{captured} in $2021+$, only $0.3\%$ (vs.\ $0.03\%$ across the whole corpus) run on $2021+$ hardware, so this
reflects the devices people carry, not the scrape date. The in-the-wild data
therefore probes only the most \emph{conservative}, pre-generative slice of the
computational-photography hypothesis; settling it needs controlled capture on
current devices (\cref{sec:conclusion}).

\paragraph{Capture conditions do not explain the false alarms.}
Since the EXIF exposure triangle is near-complete (\cref{sec:laion_mobile}), we
also stratified the false-alarm rate by physical illumination (the ISO-100
exposure value), ISO band and time of day. Most detectors are flat: ten of
twelve vary by less than $8$ percentage points across illumination bands,
including every CLIP-feature detector; the only
exceptions are AIDE ($24.6$ points higher FPR in very low light) and NPR
($12.0$). False alarms are thus driven by the processing pipeline and the
detector's training distribution, not by scene brightness.

\subsection{Threshold Calibration Cannot Save Default-Biased Detectors}
\label{sec:calibration}

\begin{figure}[tbp]
  \centering
  \includegraphics[width=\linewidth]{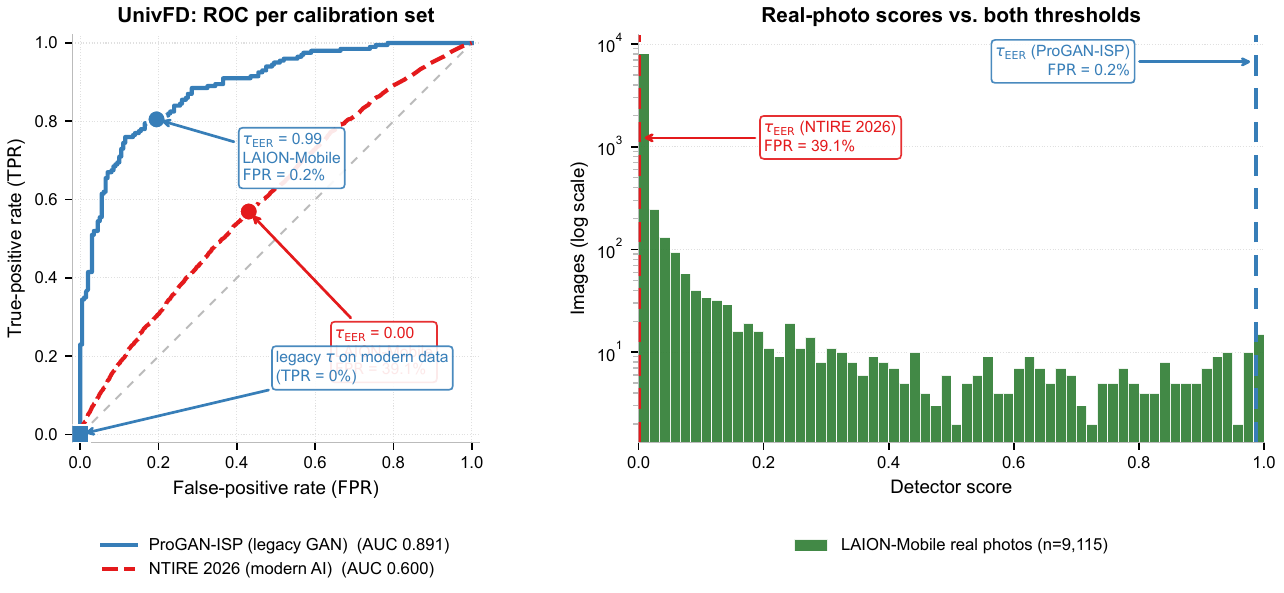}
  \caption{Calibration drift on the best-case detector (UnivFD).
  \emph{Left:} ROC curves on the two calibration sets, each EER operating point
  annotated with the LAION-Mobile false-alarm rate its threshold produces; the
  square marker projects the legacy threshold onto the modern curve, where it
  detects essentially nothing. \emph{Right:} score distribution of the real
  smartphone photos (log scale) with both EER thresholds. The two defensible
  calibrations sit at opposite ends of the score range, yielding $0.2\%$ vs.\
  $39.1\%$ FPR on identical photographs.}
  \label{fig:roc-univfd}
\end{figure}

A natural defence of the high-FPR detectors is that $\tau{=}0.5$ is
mis-calibrated. \Cref{fig:roc-univfd} makes this concrete for the best case,
UnivFD: its ROC curve is near-perfect on the legacy GAN set but sags on modern
content, so its two EER calibrations sit at opposite ends of the score range and
the legacy threshold, projected onto the modern curve, detects essentially
nothing. Moving the threshold shifts the operating point but cannot recover
performance once the score distributions overlap. Five detectors have an NTIRE
AUC whose $95\%$ CI lies entirely below $0.5$ (AUC is threshold-independent), so
no threshold helps; and the detectors that looked ``zero-FPR'' at $\tau{=}0.5$
flip to LAION-Mobile false-alarm rates above $50\%$ once their threshold is fit
on modern content (\cref{tab:laion_mobile-fpr}), confirming that apparent safety
was a calibration artefact, not a property of the detector.

\section{Discussion}
\label{sec:discussion}

\paragraph{The artefact-convergence problem.}
One reading of \cref{sec:ntire,sec:ki-era} is that the artefact-fingerprint
hypothesis behind most detectors (upsampling operator, GAN spectrum or VAE
reconstruction) has converged with the side-effects of modern computational
photography: smartphone JPEGs already contain learned-network artefacts (HDR
fusion, lens-corrected sharpening, semantic denoising), and a detector trained
to flag any such artefact may misfire on real photos. The size and even the
sign of the effect are detector-specific, and our era stratification is
correlational, not causal: the position paper's prediction that heavier ISP
processing raises the false-alarm rate does not hold for most detectors
(\cref{sec:ki-era}). That variability is itself the central observation:
behaviour on modern smartphone photography is governed less by the image's
authenticity than by the interaction between training-distribution and
capture-pipeline artefacts. Two caveats bound this. First, our corpus barely
reaches the modern ISP era, so the convergence is a \emph{conservative lower
bound} measured on first-generation ($2018$--$2020$) pipelines
(\cref{sec:ki-era}). Second, calibration drift alone already swings a single
detector between negligible and double-digit false alarms (\cref{fig:roc-univfd}),
without ever testing a current flagship. Both point the same way: adjudicating
deepfake detection on today's phones requires controlled, paired data from
current devices (\cref{sec:conclusion}).

\paragraph{Is any detector deployable?}
No. Five detectors clear chance on modern AI content, but the best reaches only
AUC $0.624$, and at a modern-calibrated threshold each still flags a large
fraction of real smartphone photos (\cref{tab:laion_mobile-fpr}).
UnivFD~\cite{ojha2023univfd} comes closest (highest discrimination, lowest
modern-calibrated false-alarm rate), yet its $39\%$ remains far too high for
content moderation, and the apparent escape hatch of calibrating on legacy GAN
data is fit to a 2019 distribution that degrades sharply on modern photographs
(\cref{sec:calibration}). The rest either invert their score ordering on modern
content or fail at every operating point (\cref{tab:laion_mobile-fpr}).

\section{Limitations}
\label{sec:limitations}

LAION-Mobile is one-class real-only by construction, so we cannot report
two-class AUC on it. Authenticity is inferred from EXIF
\texttt{Make}/\texttt{Model} plus the non-photo filtering of
\cref{sec:laion_mobile}, not verified per image: EXIF can be copied or forged,
and a web corpus may still contain edited images carrying camera EXIF. The
three-stage filter removes the bulk of mislabelled non-photographs, but the
\texttt{real} label remains a high-precision heuristic rather than certified
ground truth. Because EER thresholds are fit on the synthetic two-class sets and transferred
to LAION-Mobile, each reported FPR also reflects cross-domain score-scale
shift, the calibration drift we report. The era stratification rests on
manufacturer-disclosed specifications and may mis-classify firmware-updated
phones; more fundamentally, the corpus is dominated by pre-$2021$ hardware and
cannot probe current-flagship ISPs (\cref{sec:ki-era}). We report a single rerun per (detector, dataset); forward passes are
deterministic, but the bootstrap CIs do \emph{not} capture training-run
variance. Finally, copyright precludes redistributing the source images, so they
must be re-fetched from URLs, a fraction of which decay. We release fixed
per-image scores and content hashes, so the reported numbers stay reproducible
and re-fetched images can be verified against ours.

\section{Conclusion and Future Work}
\label{sec:conclusion}

We audit twelve detectors on their paper benchmarks, NTIRE 2026, and a
$9{,}115$-image re-LAION-5B smartphone subset. The detectors with per-generator
paper tables reproduce their published AUC, so the pipeline is sound;
generalisation then collapses on modern AI content, where only five detectors
beat chance, and once calibrated on that content even these flag $39$--$63\%$ of
real smartphone photos (\cref{tab:laion_mobile-fpr}). Real-world false-alarm
rates are thus a property less of the detector than of its calibration: the
\emph{same} detector can look deployable or unusable depending only on which
legacy set fixes its threshold. No detector in our roster is simultaneously
above chance on modern content and deployable on real photographs. This both
confirms and qualifies the position paper~\cite{keuper2026real}: its central
worry holds at scale, since detectors do flag real smartphone photographs as
fake, but the false alarms track each detector's training distribution and
threshold calibration far more than the ISP era itself (\cref{sec:ki-era}), so
computational photography is corroborated in its \emph{effect} yet, on this web
corpus, not isolated as the \emph{cause}.

A second, more fundamental gap is the device era. LAION-Mobile mirrors the
device mix of web image collections, dominated by pre-$2018$ phones
(\cref{fig:device-coverage}), so it tests only the most \emph{conservative}
version of the computational-photography argument. The sharpest test is still
ahead: today's flagships (iPhone~13 and later) make up just $0.03\%$ of our
release-dated corpus and the cutting edge is absent entirely, yet their neural
ISP fuses, denoises and re-synthesises so extensively that the boundary between
a captured and a generated pixel becomes genuinely ambiguous. We therefore plan
to collect a real-smartphone set from current devices, with paired RAW and
ISP-processed variants and image-to-image generated counterparts; it closes
the AUC-on-smartphone gap the one-class LAION-Mobile data cannot fill and
lets us test whether lightweight adaptation recovers deployable performance.

\small
\bibliographystyle{plain}
\bibliography{main}

\end{document}